\documentclass[letterpaper, 10 pt, conference]{ieeeconf}  

\IEEEoverridecommandlockouts                              

\usepackage{graphics} 
\usepackage{epsfig} 
\usepackage{amsmath} 
\usepackage{amssymb}  
\usepackage[style=ieee]{biblatex}
\usepackage{bm}
\usepackage{placeins}
\usepackage{tabularx}
\usepackage{xspace}
\usepackage{multirow}

\graphicspath{ {./images/} }

\makeatletter
\DeclareRobustCommand{\onedot}{\futurelet\@let@token\@onedot}
\def\onedot{\ifx\@let@token.\else.\null\fi\xspace}
\def\eg{\emph{e.g}\onedot}

\def\etal{\emph{et al}\onedot}
\makeatother
\newcommand{\R}{\mathbb{R}}
\newcommand{\norm}[1]{\left\lVert#1\right\rVert}

\DeclareMathOperator*{\argmin}{arg\,min}

\title{\LARGE \bf
City-scale Scene Change Detection using Point Clouds
}

\author{Zi Jian Yew$^{1}$ and Gim Hee Lee$^{1}$
\thanks{*This work is  supported in part by the Singapore MOE Tier 1 grant R-252-000-A65-114.}
\thanks{$^{1}$Zi Jian Yew and Gim Hee Lee are with Department of Computer Science, National University of Singapore
        {\tt\small \{zijian.yew, gimhee.lee\}@comp.nus.edu.sg}}%
}

\begin{document}

\maketitle
\thispagestyle{empty}
\pagestyle{empty}

\begin{abstract}

We propose a method for detecting structural changes in a city using images captured from vehicular mounted cameras over traversals at two different times. We first generate 3D point clouds for each traversal from the images and approximate GNSS/INS readings using Structure-from-Motion (SfM). A direct comparison of the two point clouds for change detection is not ideal due to inaccurate geo-location information and possible drifts in the SfM. To circumvent this problem, we propose a deep learning-based non-rigid registration on the point clouds which allows us to compare the point clouds for structural change detection in the scene. Furthermore, we introduce a dual thresholding check and post-processing step to enhance the robustness of our method. We collect two datasets for the evaluation of our approach. Experiments show that our method is able to detect scene changes effectively, even in the presence of viewpoint and illumination differences.

\end{abstract}

\section{INTRODUCTION}


3D point clouds reconstructed from image-based Structure-from-Motion (SfM) are often frozen in time and thus gradually loses its ability to model the constantly changing environment with high fidelity. The first step towards maintaining an up-to-date city-scale 3D model is to detect changes 
in the geometric structure of the scene, while excluding other nuisance factors such as appearance changes from illumination or viewpoint differences. Detecting temporal changes in a city is an important problem, with many applications such as maintaining updated maps for autonomous driving systems \cite{learnedDeconv}, surveillance \cite{Sidike2015cdsurveillance}, and disaster damage assessment \cite{sakurada13}. 

One na\"ive way of computing the changes is to directly compare images between the two traversals using some variant of image differencing \cite{imagediff,changedettechniques}. However, such approaches are sensitive to illumination differences between the two acquisitions. In addition, they require near pixel-perfect alignment to work well which can be hard to achieve on a moving camera. Several works \cite{learnedDeconv,taneja2011} tackle this problem using a 3D model of the scene. These typically reconstruct a dense 3D model and recover the camera poses using SfM and multi-view stereo (MVS) techniques. The dense models can later be used for dense image alignment \cite{learnedDeconv} or to reproject pixels between images to detect inconsistencies \cite{taneja2011}. However, obtaining accurate camera poses across traversals can be challenging. GNSS errors can often be several meters in urban environments; SfM techniques can also fail due to appearance differences which can be due to illumination differences or even large scene changes. In addition, reconstructing the dense model is computationally expensive and dense reconstructions are arguably unnecessary for many applications such as localization of autonomous vehicles \cite{shi2019visloc}. 

To avoid these problems, we register and detect changes using the 3D point clouds instead of registering images across traversals. We first reconstruct sparse point clouds separately for each traversal using SfM \cite{schoenberger2016sfm}. The reconstruction can be fairly robust since each traversal is captured within a short timespan and contains limited scene and illumination changes. The reconstructed point clouds from the two traversals are geo-registered using GNSS/INS data, but may not align perfectly due to GNSS/INS and reconstruction inaccuracies. We tackle this by performing a non-rigid registration to warp one of the point clouds to the other.

\begin{figure}[t]
\begin{center}
\includegraphics[width=\linewidth]{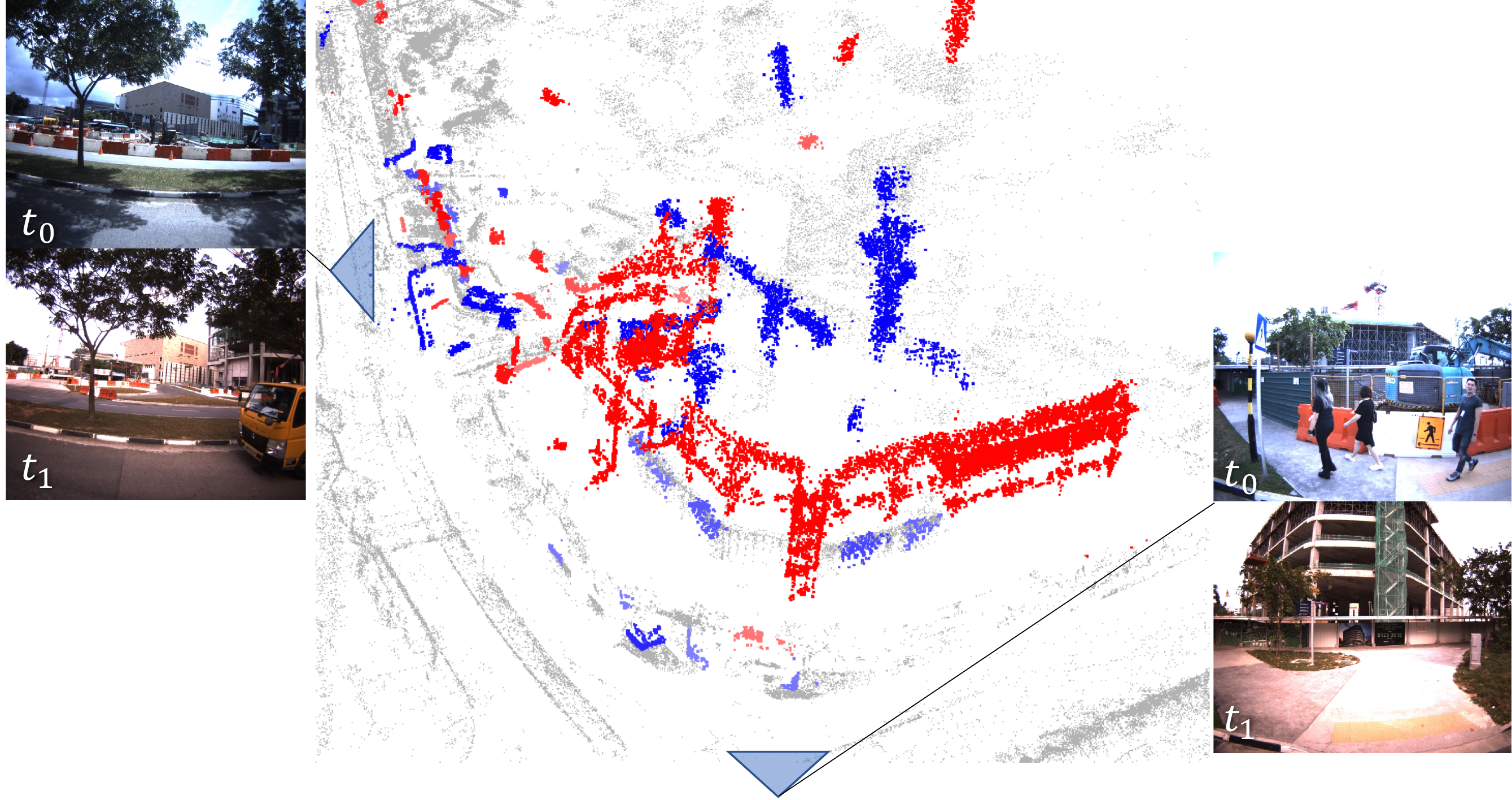}
\end{center}
\vspace{-3mm}
\caption{Visualization of changes detected in the Business District using our approach. Blue and red indicate points which disappeared or appeared respectively (i.e. only present during $t_0$ or $t_1$). We also show images capturing the changed scene. Our approach detects the appearance of a new building as well as the disappearance of cranes and road barriers.}
\label{fig:qualitative3d-teaser}
\end{figure}

The changes can now be detected by comparing the two point clouds. Comparing point clouds generated from SfM comes with its own challenges. In particular, reconstructions of the same scene at different times may vary significantly due to variations in imaging conditions. To alleviate this issue, we employ a dual thresholding scheme where we compare between subsampled and original point clouds to detect changes. The point clouds are subsampled by considering only points which can be reliably observed from a larger number of images. These points are more stable and likely to be reconstructed in the other traversal.

We demonstrate the effectiveness of our approach on two datasets collected over two different areas. Each dataset contains images and GNSS/INS readings of the same route over two traversals that are roughly two months apart. Our datasets contain large changes \eg building construction or demolition, and smaller ones, \eg vehicle movements or tree planting. Experiments show our approach can detect these structural changes even in the presence of viewpoint and illumination differences. Fig. \ref{fig:qualitative3d-teaser} shows an example of our detection result.
Our contributions are as follows:
\begin{itemize}
    \item Propose a deep learning-based non-rigid point cloud registration to align two imperfect point clouds.
    \item Design a point cloud comparison scheme to reliably detect changes.
    \item Collect two datasets of images and GNSS/INS readings to validate the effectiveness of our approach. Our reconstructed point clouds and annotated image pairs will be made available on our project webpage\footnote{\url{https://yewzijian.github.io/ChangeDet}}.
\end{itemize}

\vspace{1mm}
\section{RELATED WORK}
\noindent \textbf{Change Detection.} Change detection methods can be broadly classified as 2D or 3D. 2D methods generally compare input images though some variant of image differencing \cite{imagediff,changedettechniques}, and tend to be sensitive to illumination differences or misalignments between the images. To partially overcome these issues, the images are often preprocessed to remove illumination variations \cite{radke2005survey}, and registered \cite{regtechniques} to each other. Despite this, 2D methods remain sensitive to viewpoint changes as they typically require pixel perfect registration to work well. Also, most 2D methods detect appearance changes which may not correspond to actual scene changes.
3D methods make use of a known 3D scene structure or reconstruct it from the input images to better detect structural changes. 
Taneja \etal \cite{taneja2011} assumes that a 3D model of the scene in the previous time step is available and detects changes by checking the consistency during projection between images in the later time step.
Ulusoy and Mundy \cite{ulusoy2014image} extends this to infer changes in the 3D model itself. Sakurada \etal \cite{sakurada13} foregoes the dense model and instead makes use of stereo pairs in both time steps to perform the reprojection.
Another direction is to generate a spatio-temporal model from images captured at various times by incorporating time into SfM methods.  \cite{Schindler_Dellaert_2010,Matzen_Snavely_2014} infer the temporal ordering of images and the temporal extent of 3D points in the scene by analyzing the SfM output. Lee and Fowlkes \cite{Lee_Fowlkes_2017} optimizes a probabilistic spatial-temporal model using expectation-maximization to simultaneously register 3D maps and infer the temporal extents of scene surfaces.
Most of the above methods require accurate relative camera poses between the two times, which can be difficult to obtain especially when the scene has changed.

\vspace{1mm}
\noindent \textbf{Change Detection using Deep Learning.} More recently, several works \cite{learnedDeconv,guo2018learning,sakurada2020weakly,stent2015,siameseContrastive2017,changenet2018} apply deep learning to detect scene changes. These works learn to compare two input images to detect changes in a strongly supervised manner, requiring pixel level \cite{learnedDeconv,siameseContrastive2017,changenet2018,guo2018learning} or patch level \cite{stent2015} annotations which can be highly tedious to obtain.
To avoid the need for annotation, Sakurada and Okatani \cite{sakurada2015cnnfeat} compare normalized features extracted from an upper layer of a convolutional neural network pretrained on a image recognition task, and make use of super-pixel segmentations to obtain high resolution outputs.
Despite the good performance shown by these works, these image-based methods remain sensitive to viewpoint differences since they do not consider the 3D structure of the scene. Alcantarilla \etal \cite{learnedDeconv} partly alleviates this issue by performing a dense warp between images, but requires a computationally expensive dense reconstruction and accurate relative camera poses across the two traversals.

\vspace{1mm}
\noindent \textbf{Point Cloud Registration.} The above change detection works often require accurate image registration, which can be difficult to achieve under scene changes or illumination variations. We circumvent these challenges by generating 3D point clouds from the input and registering the point clouds instead. Point cloud registration methods can be broadly classified into 1) feature-based methods \cite{FPFH,USC,SHOT} which establish correspondences by matching descriptors before computing the transformation, and 2) simultaneous pose and correspondence methods \cite{icp,chen-medioni} that typically use iterative schemes to estimate both pose and correspondences. Learned variants of both feature-based \cite{3dmatch,cgf,ppfnet,3dfeatnet,fcgf} and simultaneous pose and correspondence \cite{pointnetlk,deepmapping,deepcp,RPMNet} methods are also available. One particular work, DeepMapping \cite{deepmapping} optimizes for the registration objective by training a neural network. Depending on the application, point clouds may undergo local deformations which requires estimating a non-rigid transformation. In this work, we extend DeepMapping to handle non-rigid deformations through the use of Gaussian Radial Basis Functions (RBFs), which have been used in many point cloud registration works \cite{PointSetReg-GMM,coherentPtDrift,Chui-RPM}.

\vspace{1mm}
\section{Our Change Detection Pipeline: Overview}
\begin{figure}[ht]
\begin{center}
\includegraphics[width=\linewidth]{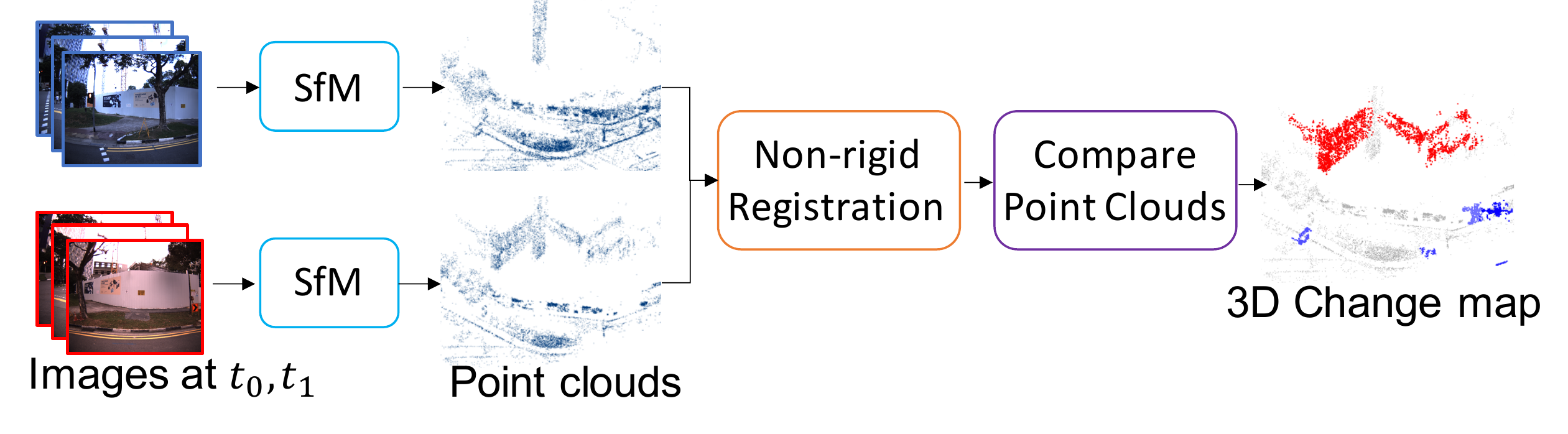}
\end{center}
\vspace{-5mm}
\caption{Our change detection pipeline}
\vspace{-3mm}
\label{fig:whole-pipeline}
\end{figure}

Fig. \ref{fig:whole-pipeline} shows our change detection pipeline.
The inputs are images from two time steps and their approximate GNSS/INS poses (Section \ref{sect:data-acquisition}). We first use SfM to reconstruct geo-registered point clouds (Section \ref{sect:sfm}). Inaccuracies in SfM result in deformations of the reconstructed point clouds. We remove these deformations by applying a non-rigid registration to align the two point clouds (Section \ref{sect:non-rigid-reg}). Finally, the registered point clouds are compared to detect the scene changes (Section \ref{sect:compare-clouds}). For convenience, we list important algorithm parameters in Table \ref{table:parameters}.

\begin{table}[ht]
\small
\begin{center}
\caption{Table of parameters}
\label{table:parameters}
\begin{tabularx}{0.9\linewidth}{X | c c}
  \hline
  Module & Parameter & Value \\
  \hline
  \multirow{3}{*}{Registration} & \# basis points, $K$ & 36 \\
  & Max. chamfer distance $\delta_{reg}$ & $10$ \\
  & Regularization weight, $\lambda_{reg}$ & 0.01 \\
  \hline
  \multirow{4}{2cm}{Change detection} & Subsampling track length $\tau_{ss}$ & 7\\ 
   & Max. chamfer distance $\delta_{cd}$ & 10 \\
   & \# neighbors for mean filtering, $k$ & 7 \\
   & \# Min. distance for changes, $\tau_{cd}$ & 2.0 \\
  \hline
\end{tabularx}
\vspace{-2mm}
\end{center}
\end{table}

\section{FROM IMAGES TO POINT CLOUDS}

\subsection{Data Acquisition}\label{sect:data-acquisition}
Our data acquisition platform is a vehicle with two side-mounted wide-angle color cameras. The cameras capture images with a resolution of $2464 \times 2056$, and are set to auto-exposure to adapt to different lighting conditions. The vehicle is equipped with a GNSS/INS system which is time-synchronized to the camera system.
Images are captured every 0.6m of distance traveled. This distance was chosen taking into account reconstruction efficiency, while still allowing nearby objects to be observed from multiple images.

\subsection{Sparse Reconstruction}\label{sect:sfm}
\begin{figure}[ht]
\begin{center}
\includegraphics[width=\linewidth,trim={0 0.7cm 0 0.0cm},clip]{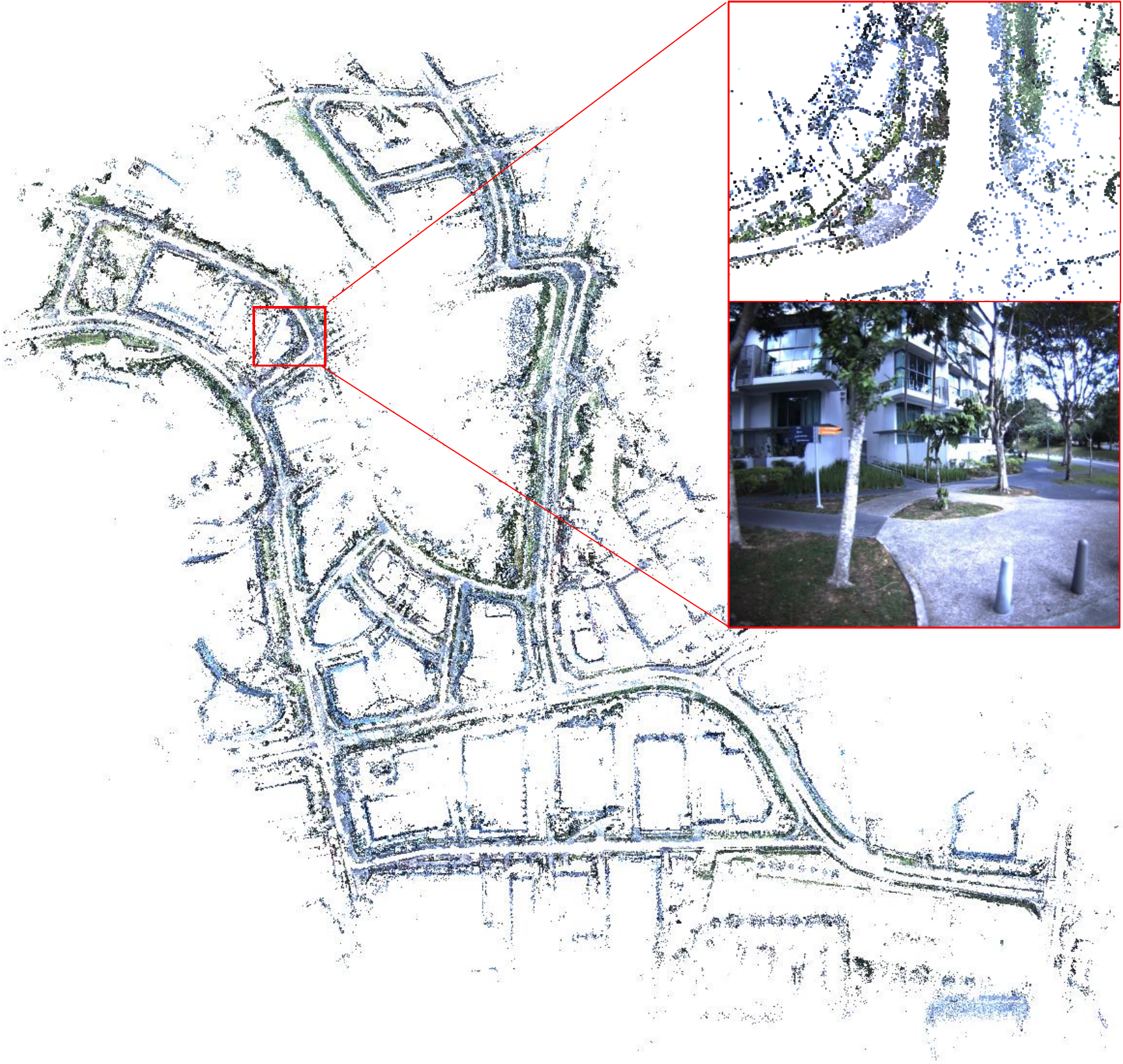}
\end{center}
\vspace{-3mm}
\caption{Our reconstructed point cloud of the Business District (BD) dataset. We also show a zoomed in view of the point cloud and a sample image.}
\label{fig:sfm-examples}
\end{figure}

We generate sparse point clouds from the images using a modified version of the COLMAP \cite{schoenberger2016sfm} SfM pipeline. Following \cite{autovision}, we minimize the time required for reconstruction by only matching images that are within 20m, and use the GNSS/INS readings to initialize the camera poses. Fig. \ref{fig:sfm-examples} shows an example of the reconstructed point cloud with our Business District (BD) dataset.

\section{DETECTION OF CHANGES}

\subsection{Non-rigid Registration}\label{sect:non-rigid-reg}
The outputs from the previous stage are two geo-registered point clouds. However, GNSS/INS and reconstruction inaccuracies lead to local deformations in the reconstructed point clouds. The local deformations result in misalignment between the two point clouds, and they cannot be compared directly for change detection. To alleviate this problem, we perform non-rigid registration to reduce the misalignment of the two point clouds. More formally, given a reference $\mathbf{P}_{ref}$ and source $\mathbf{P}_{src}$ point clouds, the goal of non-rigid registration is to find the parameters $\bm{\theta}$ of the non-rigid warping $T(\mathbf{P}_{src}; \bm{\theta})$ that warps $\mathbf{P}_{src}$ into the best alignment with $\mathbf{P}_{ref}$, i.e.
\begin{equation}
    \argmin_{\bm{\theta}} \mathcal{L}(\mathbf{P}_{ref}, T(\mathbf{P}_{src}; \bm{\theta})). \label{Eq:totalLoss}
\end{equation}
We parameterize the non-rigid warping $T(.)$ using Gaussian Radial Basis Functions (RBFs). The quality of alignment between reference and warped source point clouds $\mathcal{L}$ is given by the squared Chamfer distance, and a regularizer on the parameters. Instead of a direct minimization of $\mathcal{L}$ using solvers such as 
Levenberg–Marquardt, we use it as the registration loss in a deep neural network that takes $\mathbf{P}_{ref}$ and $\mathbf{P}_{src}$ as inputs and outputs the optimal $\bm{\theta}$. 
We now describe these components in more detail.

\subsubsection{Gaussian RBF model}
Gaussian Radial Basis Functions (RBFs) have been used for warping images \cite{imagewarp} and point clouds \cite{PointSetReg-GMM, coherentPtDrift} due to its implicit smoothness. The smoothness property is important as it discourages warpings which are too arbitrary.
A Gaussian RBF with $K$ basis points maps a 3D point $\mathbf{x}_i \in \R^3$ to its target position $T(\mathbf{x}_i; \bm{\theta}) \in \R^3$ with the following function\footnote{The equation assumes no rigid/affine component for the warp, which is reasonable since the point clouds are already registered using GNSS/INS.}:
\begin{equation}
    T(\mathbf{x}_i; \bm{\theta}) = \mathbf{x}_i + \phi(\mathbf{x}_i) \cdot \mathbf{W},
\end{equation}
where $\mathbf{W}$ is the $K \times 3$ warping coefficient matrix. $\phi(x_i)$ is the RBF kernel, and is a $1 \times K$ vector for each point $\mathbf{x}_i$:
\begin{equation}
\phi(\mathbf{x}_i)= 
\begin{bmatrix}
g(\norm{\mathbf{x}_i - \mathbf{c}_{1}}) & \dots & g(\norm{\mathbf{x}_i - \mathbf{c}_{K}})
\end{bmatrix},
\end{equation}
where $\mathbf{c}_{1...K}$ denotes the anchor centers for the warp, $\norm{\cdot}$ denotes the $\ell^2$ norm and
the kernel function is a Gaussian form, i.e. $g(t) = e^{-t^2/\sigma^2}$.
Intuitively, the anchor centers $\mathbf{c}_i$'s control the regions to warp, and $\mathbf{W}$ controls the magnitude and direction of the warp around each of these regions. Since the 3D point clouds are reconstructed from images captured from a car moving on the ground, drifts typically occur along the $x$ and $y$ directions. Therefore, we only consider the distances along the $xy$ plane when computing the kernel values. Putting the Gaussian RBF into Eq.~\ref{Eq:totalLoss}, the goal now becomes finding the optimal RBF parameters $\bm{\theta} = \{\mathbf{c}_1, ..., \mathbf{c}_K, \sigma_i, ..., \sigma_K, \mathbf{W}\}$. Assuming spherical covariances for the Gaussians, there are a total of $2K + K + 3K = 6K$ parameters to be estimated.

\subsubsection{Registration Loss $\mathcal{L}$}\label{sect:regloss}
To encourage the alignment of the points, we minimize the squared Chamfer distance between the reference $\mathbf{P}_{ref}$ and the transformed source $\mathbf{P}_{src}' = T(\mathbf{x}_i; \bm{\theta})$ point clouds. The squared Chamfer distance between two point clouds $\mathbf{X, Y}$ is defined as:
\begin{equation}
\begin{split}
    \mathcal{L}_{CD}(\mathbf{X}, \mathbf{Y}) = 
      &
      \frac{1}{|\mathbf{X}|} \sum_{\mathbf{x} \in \mathbf{X}}{\rho_{reg}(\min_{\mathbf{y} \in \mathbf{Y}}{\norm{\mathbf{x-y}}}^2)} \\
      & + 
      \frac{1}{|\mathbf{Y}|} \sum_{\mathbf{y} \in \mathbf{Y}}{\rho_{reg}(\min_{\mathbf{x} \in \mathbf{X}}{\norm{\mathbf{x-y}}}^2)},
\end{split}
\end{equation}
where $\rho_{reg}(\cdot)= \min(\cdot, \delta_{reg})$ clamps the Chamfer distance for robustness, and we set $\delta_{reg}=10 \text{m}^2$ in all experiments.

Despite the use of smooth Gaussian RBFs, we observe empirically that severe warping may still occur, particularly in regions with changes. We further encourage smoothness by introducing an additional term $\mathcal{L}_{reg}$ to regularize the warping by penalizing large motions of small regions:
\begin{equation}
    \mathcal{L}_{reg}(\mathbf{W}, \sigma_{1, ..., K}) = \frac{1}{K} \sum_{i = 1}^{K}{\frac{\norm{\mathbf{w}_i}}{\sigma_i^2}},
\end{equation}
where $\mathbf{w}_i$ denotes the $i^\text{th}$ row of $\mathbf{W}$.
The final loss is then the sum of the two losses:
\begin{equation}
\label{eq:total-loss}
    \mathcal{L} = \mathcal{L}_{CD} + \lambda_{reg} \mathcal{L}_{reg}.
\end{equation}

\subsubsection{Optimization with Neural Network}
\begin{figure}[ht]
\begin{center}
\includegraphics[width=\linewidth]{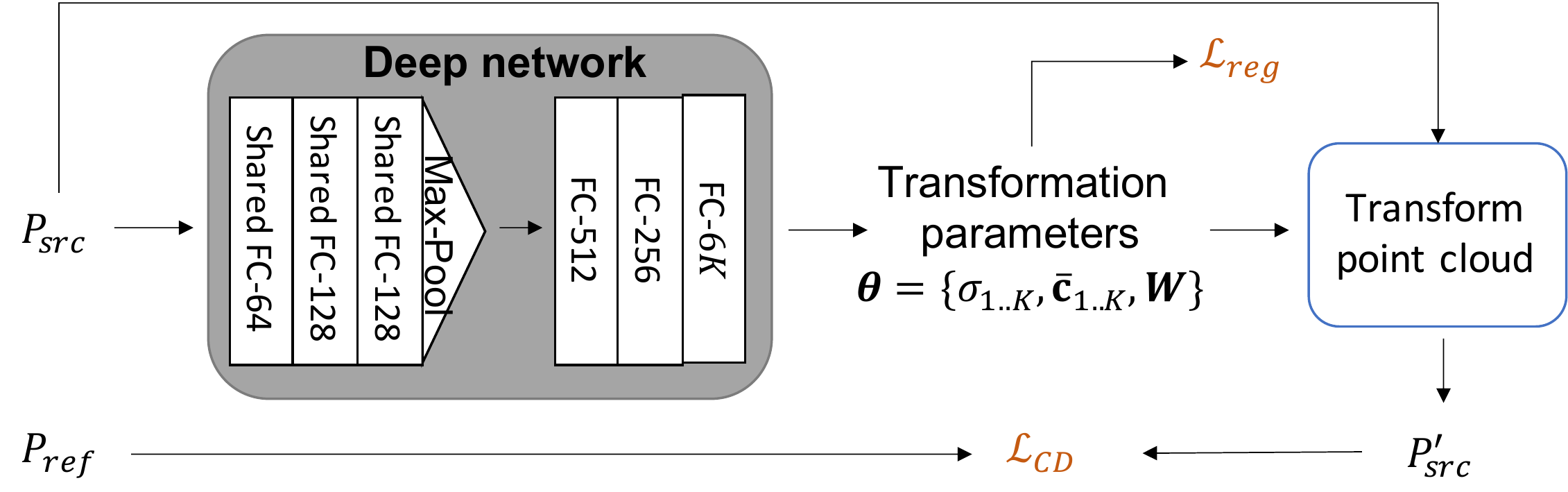}
\end{center}
\vspace{-3mm}
\caption{Our network architecture for non-rigid registration (FC-$N$: denotes fully-connected layer with $N$ output nodes).}
\label{fig:network}
\end{figure}

Inspired by DeepMapping \cite{deepmapping}, we design a deep neural network for the non-rigid registration.
As explained in \cite{deepmapping}, an indirect optimization using a deep network is akin to changing variables \cite{agrawal2018rewriting} and leads to an empiricially easier optimization.
Our network is shown in Fig. \ref{fig:network}. We feed $\mathbf{P}_{src}$ into a permutation invariant PointNet \cite{pointnet} network, which outputs the transformation parameters $\bm{\theta}$ for alignment to $\mathbf{P}_{ref}$.
We omit activations for the outputs with the exception of $\sigma_{1..K}$, where we use the softplus activation to constrain them to be positive. Furthermore, instead of predicting the absolute positions for the Gaussian centers $\mathbf{c}_{1..K}$, we find it beneficial to predict the offsets from predefined positions sampled using a uniform 2D grid over the entire point cloud. 
To register the point clouds, we simply train the network to reduce the alignment and regularization losses in Eq. \ref{eq:total-loss} using the Adam optimizer \cite{kingma2015adam} with a learning rate of 5e-4. For all experiments, we set $K=6^2=36$.

\subsection{Detecting Changes by Comparing Point Clouds}\label{sect:compare-clouds}
\begin{figure}[ht]
\begin{center}
\includegraphics[width=\linewidth]{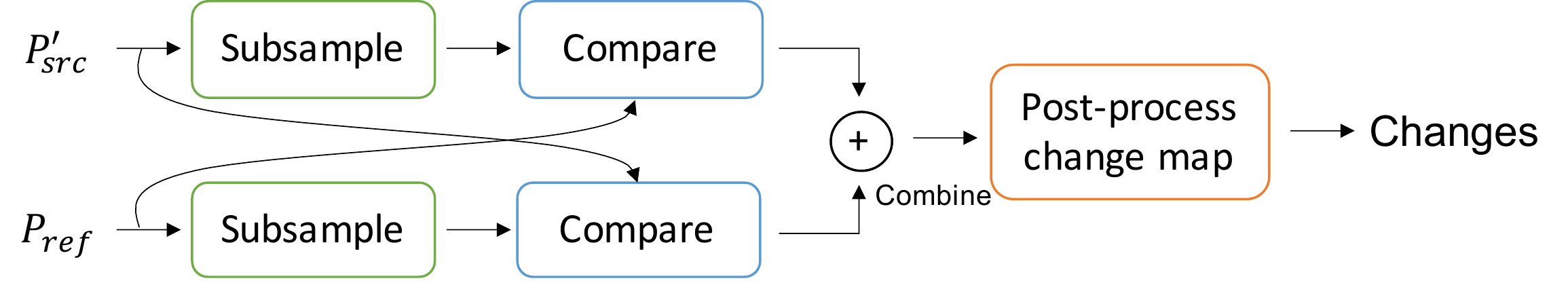}
\end{center}
\vspace{-3mm}
\caption{Point cloud comparison pipeline}
\label{fig:cd-pipeline}
\end{figure}

The reference and transformed source point clouds are compared in this step to compute the changed regions.
Fig. \ref{fig:cd-pipeline} illustrates our change detection pipeline. Point clouds reconstructed through SfM contain a fair amount of noise and variation, and thus our change detection pipeline compares point clouds using a dual thresholding scheme and applies filtering to clean up the change map. 

\subsubsection{Dual Thresholding}\label{sect:dualthreshold}
Points reconstructed from SfM may vary between reconstructions due to many factors (\eg illumination), which lead to false positives during comparison. To circumvent this problem, we employ a \emph{dual thresholding} scheme, where two different sampling ratios or ``thresholds'' are used to detect the disappearance and appearance of points.
Specifically, when detecting new points in the transformed source point cloud ${\mathbf{P}}_{src}'$, we compare its \emph{subsampled} version with the \emph{original} reference point cloud ${\mathbf{P}}_{ref}$. Similarly, we consider the subsampled ${\mathbf{P}}_{ref}$ when detecting new points in it. The points are subsampled by retaining only those with a track length above $\tau_{ss}=7$. The motivation is that these points are observable in large number of images with different viewpoints, and are likely to be more stable and reconstructed in the other point cloud. We also detect and remove points on the ground during this stage as we find them to be less repeatable between reconstructions.

\subsubsection{Point Cloud Comparison}
We compute the change response by considering the distances between points in the two point clouds. Specifically, the change response $C(\mathbf{x})$ of a point $\mathbf{x} \in \mathbf{P}_{src}'$ describes how likely it is a changed point not present in $\mathbf{P}_{ref}$ and is given as the distance to the closest point in $\mathbf{P}_{ref}$:
\begin{equation}
    C(\mathbf{x}) = \rho_{cd} (\min_{\mathbf{y} \in \mathbf{P}_{ref}}{\norm{\mathbf{x-y}}}),
\end{equation}
where $\rho_{cd}(\cdot) = \min(\cdot, \delta_{cd})$ clamps the distance for robustness, and we set $\delta_{cd}=10$m. To increase sensitivity, we only consider point pairs with similar normals within $40^{\circ}$ of each other. Points in $\mathbf{P}_{ref}$ which are not present in $\mathbf{P}_{src}$ can be detected using a similar formulation, but this time considering the downsampled $\mathbf{P}_{ref}$.

\subsubsection{Post-processing of Change Map}
\begin{figure}[ht]
\begin{center}
\includegraphics[width=0.8\linewidth]{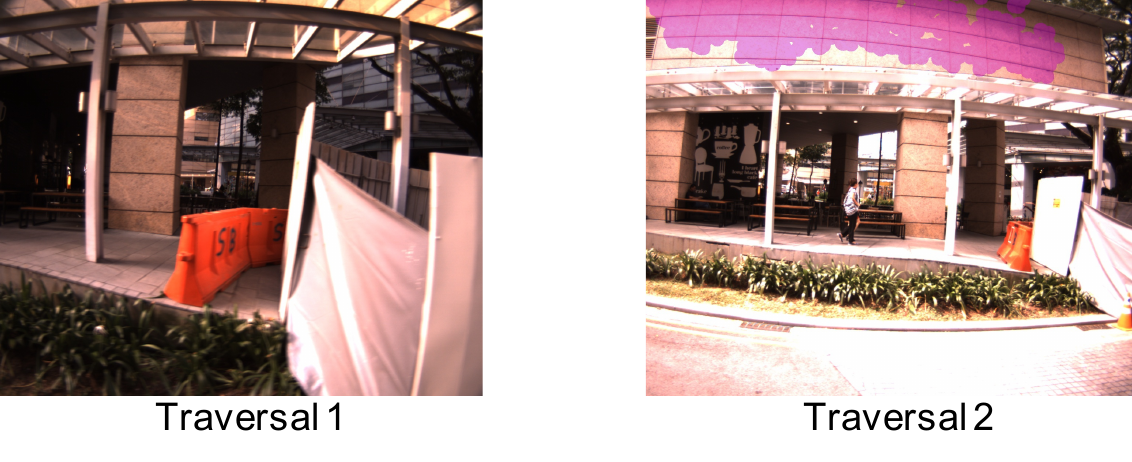}
\end{center}
\vspace{-5mm}
\caption{The camera does not observe the top of the building in the first traversal. Without considering the camera FOVs, such regions (highlighted in magenta) will be erroneously detected as changed regions.}
\label{fig:fov-check}
\end{figure}

We post-process the change map in this step to filter out spurious changes. 
We first smooth out the change responses by applying a kNN-based mean filter to improve robustness against small misalignments: for each point $\mathbf{x}$, we replace its change response $C(\mathbf{x})$ by the average response of itself and its $k=7$ nearest neighbors. Points with low change responses below $\tau_{cd}=2.0$m after filtering, as well as isolated points with few neighboring points having large change responses are removed. Subsequently, we filter out points which are not observed in the field of views (FOVs) of the cameras in the other traversal. This step is important to handle situations where points are missing simply because the camera trajectory has changed (Fig. \ref{fig:fov-check}). Lastly, we repopulate the changed points with nearby points from the point cloud before the downsampling step in Section \ref{sect:dualthreshold}.

\section{EXPERIMENTAL RESULTS}
We collect two datasets from a Business District (BD) and Research Town (RT), respectively to test our approach. Each dataset contains data from two traversals that are two months apart and cover roughly 0.3 square kilometers. BD contains many high rise buildings, while RT contains mostly low rise buildings. Despite being captured just two months apart, construction and demolition activities lead to several large structural changes in the scene. Additionally, smaller changes such as planting/removal of trees, and movement of vehicles and cranes are present in the dataset.
We randomly select a total of 30 image pairs that contain changes and annotate pixel-level structural changes between them.

\subsection{Non-rigid registration}
\begin{figure}[ht]
\begin{center}
\includegraphics[width=0.97\linewidth]{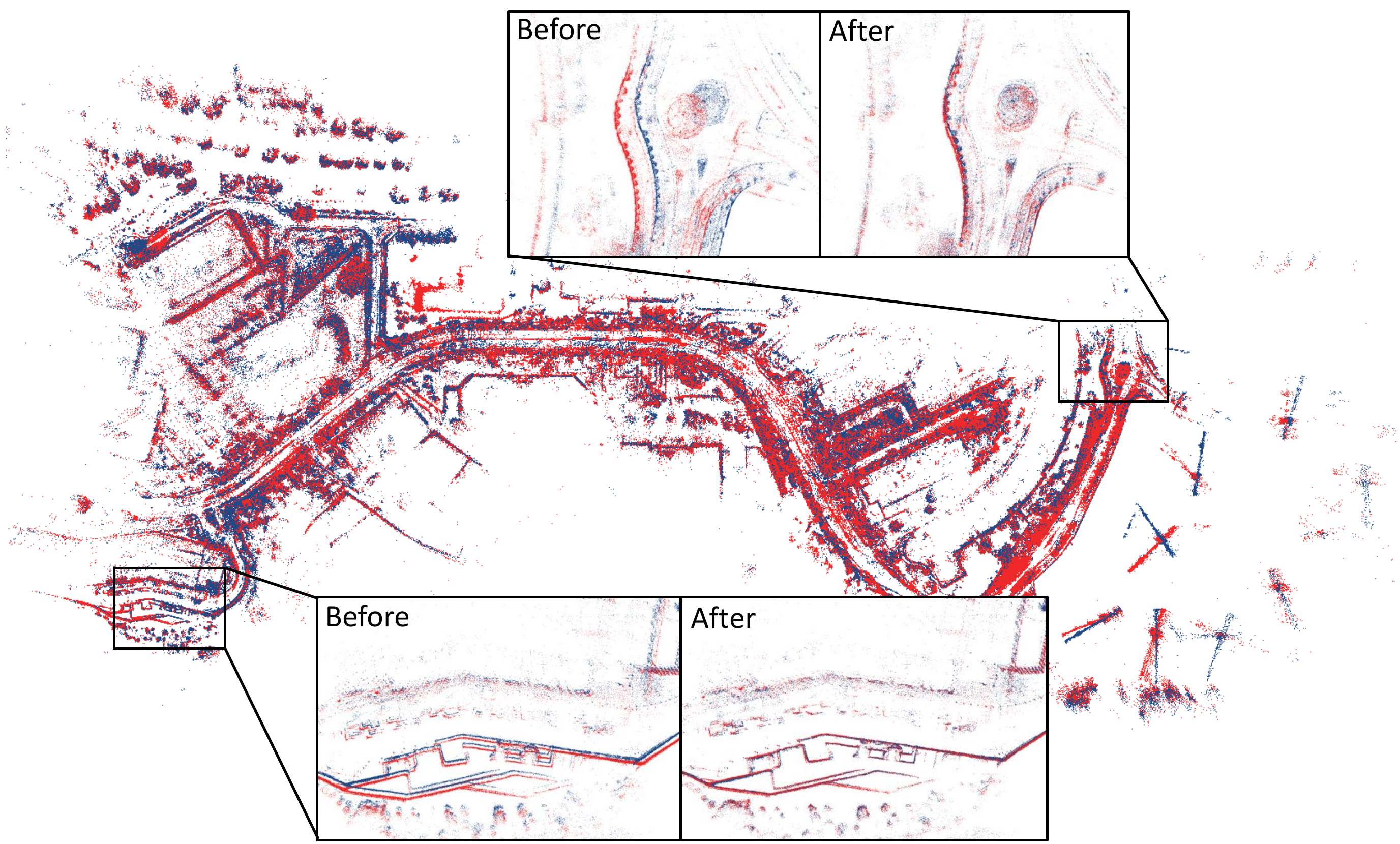}
\end{center}
\vspace{-3mm}
\caption{Qualitative results of non-rigid registration output (Research Town). We also show zoomed-in views of the point clouds before and after registration. Best viewed in color.}
\label{fig:nonrigid-outputs}
\end{figure}

\begin{figure}[ht]
\begin{center}
\includegraphics[width=0.95\linewidth,trim={0 0.3cm 0 0.3cm},clip]{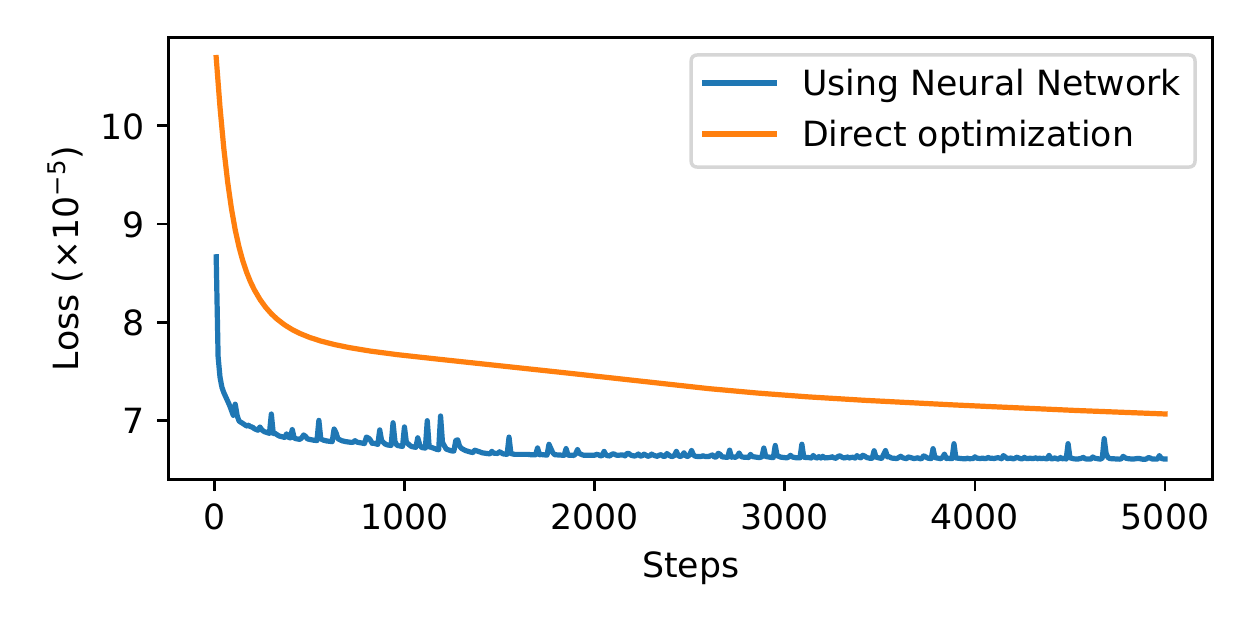}
\end{center}
\vspace{-4mm}
\caption{Registration loss during optimization. Indirect optimization with a neural network mostly converges after about 1500 steps. In contrast, direct optimization of registration parameters using the same optimizer gives a higher loss even after 5000 steps.}
\label{fig:nonrigid-loss}
\end{figure}

We first show in Fig. \ref{fig:nonrigid-outputs} the output of our non-rigid registration algorithm. We observe that the alignment of buildings and roads in the scene improve significantly after registration.
In Fig. \ref{fig:nonrigid-loss}, we also show the effectiveness of using neural networks for optimizing for the non-rigid parameters $\bm{\theta}$. Despite using the same momentum-based optimizer \cite{kingma2015adam} and learning rate, the neural-network-based optimization converges in a smaller number of steps to a better minima compared to direct optimization.

\subsection{Change Detection}
\begin{figure}[t]
\begin{center}
\includegraphics[width=\linewidth,trim={0 0 0 3cm},clip]{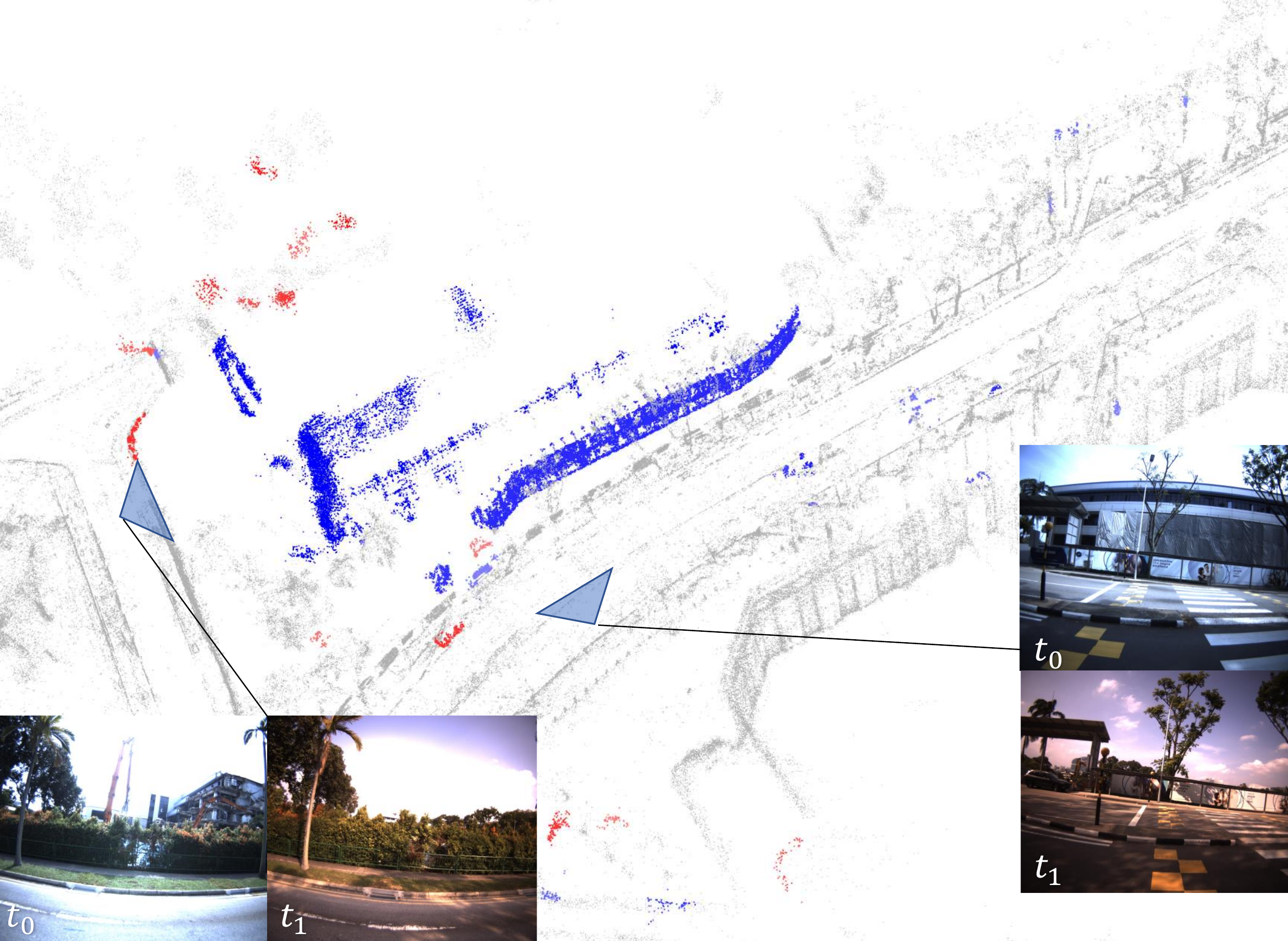}
\end{center}
\vspace{-2mm}
\caption{Visualization of the changes detected on the point clouds (Research Town). Blue and red indicate points which disappeared or appeared respectively. We also show images capturing the changed scene.}
\vspace{-2mm}
\label{fig:qualitative3d}
\end{figure}

Figs. \ref{fig:qualitative3d-teaser} and \ref{fig:qualitative3d} show qualitative examples of our change detection output. Our algorithm outputs sparse 3D point clouds that show the changed locations, and distinguishes between the disappearance and appearance of scene structures. Compared to algorithms which show the changes on images, our point cloud outputs are useful in giving a quick overview of changed regions in large scenes.

We also compare with two recent deep learning-based change detection works \cite{guo2018learning,sakurada2020weakly}. We initialize the network weights with the weights trained\footnote{Pre-trained weights are provided by the authors for \cite{guo2018learning}. For \cite{sakurada2020weakly}, we pre-train the network using the provided training code.} on the publicly available TSUNAMI dataset \cite{sakurada2015cnnfeat}. We then finetune the network weights on our dataset using a 5-fold cross validation scheme, where each fold uses a 18/6/6 train/val/test split. We evaluate the 2D change maps using the mean intersection-over-union (mIOU) as suggested in \cite{learnedDeconv,sakurada2020weakly}.
Since the evaluation is performed on 2D images, we generate 2D change maps for our algorithm by projecting detected changed points within 100m of the camera location onto the image plane, and label all pixels within a radius of 20 pixels around each projected point as changed. Note that since we do not compute dense models, our approach is unable to account for occlusion of detected changes in the images, and this may lead to additional false positives during evaluation. 

\begin{figure*}[ht]
\begin{center}
\includegraphics[width=1.0\linewidth]{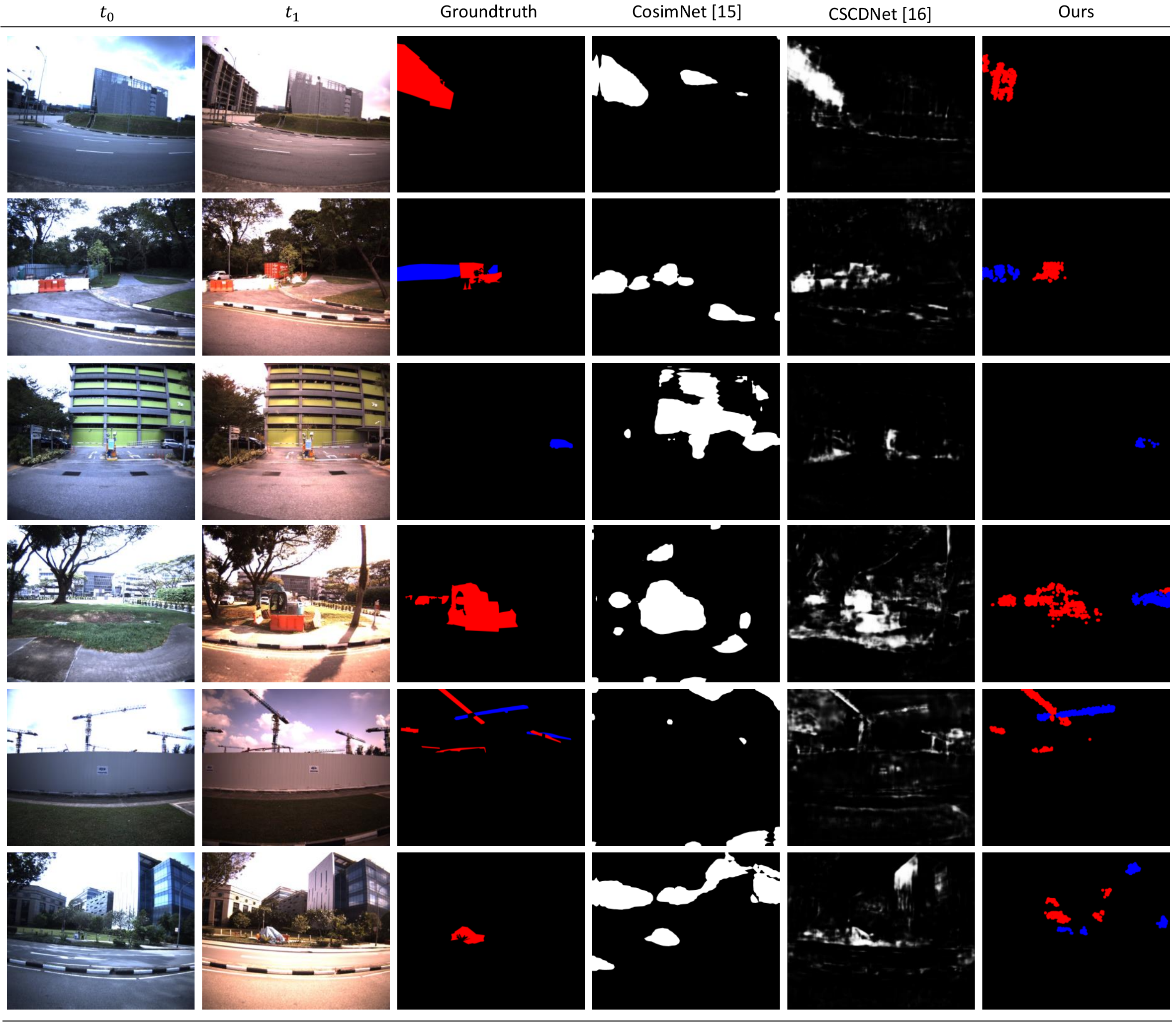}
\end{center}
\vspace{-2mm}
\caption{Examples of change detection results, and images from the two traversals. For the groundtruth and our algorithm, blue and red denote the disappearance and appearance of scene content from time $t_0 \to t_1$. The image-based baseline algorithms \cite{guo2018learning,sakurada2020weakly} are unable to differentiate between appearance/disappearance and all detected changes are drawn in white. Rows 1-3 are from the business district, and row 4-6 are from the research town. The last row shows a failure case by our algorithm. Best viewed in color.}
\vspace{-2mm}
\label{fig:qualitative2d}
\end{figure*}

\begin{table}[ht]
\small
\begin{center}
\caption{mIOU metrics on our dataset}
\label{table:cd-quantitative}
\begin{tabularx}{0.8\linewidth}{X | c c c}
  \hline
  Method & BD & RT & (All) \\
  \hline
  CosimNet \cite{guo2018learning} & 0.570 & 0.582 & 0.576 \\
  CSCDNet \cite{sakurada2020weakly} & 0.589 & 0.542 & 0.567 \\
  \hline
  Ours & \textbf{0.597} & \textbf{0.633} & \textbf{0.613} \\
  \hline
\end{tabularx}
\vspace{-2mm}
\end{center}
\end{table}

The results are shown in Table \ref{table:cd-quantitative}. We also show qualitative comparisons of our detected changes in Fig. \ref{fig:qualitative2d}. 
The viewpoint differences in our dataset makes it highly challenging for the image-based change detection algorithms. In contrast, our algorithm compares 3D point clouds and can better handle viewpoint differences. As a result, our approach outperforms image-based algorithms despite not accounting for occlusions during image projection. The last row in Fig. \ref{fig:qualitative2d} shows a typical failure case of our algorithm, where strong lighting differences (\eg in the building regions) as well as the complex foliage structures lead to inconsistent reconstructions and subsequently false alarms.

\section{Limitations of Our Approach}
Although our approach outperformed existing image-based algorithms, it currently suffers from several limitations. 
Our approach requires point clouds of both time steps to be reconstructed which can be time consuming in larger scenes, although the subsequent registration and change detection steps can be performed reasonably fast. For example, our BD dataset requires 3 days, 25min, and 2min respectively for the reconstruction, registration, and change detection steps.
Furthermore, since our algorithm operates on the output of SfM, structures which are hard to reconstruct, \eg foliage lead to many false positives. Smaller scene structures or objects without distinct features are also likely to be missed as they tend to be not well reconstructed in the point cloud.

\section{CONCLUSIONS}
We propose a workflow to perform detect structural changes in large scenes by comparing sparse point clouds. This is useful for applications where sparse point clouds are available, \eg maps generated for self driving cars, and our algorithm can detect regions in the point clouds that need to be updated. Experiments on two datasets with viewpoint differences show the feasibility of our approach.








\clearpage
\printbibliography

\end{document}